\documentclass[letterpaper, 10 pt, conference]{ieeeconf}  

\IEEEoverridecommandlockouts                              

\usepackage{graphics} 
\usepackage{amsmath} 
\usepackage{amssymb}  
\usepackage{url}
\usepackage{hyperref}
\usepackage{cite}
\usepackage{graphicx}
\usepackage{multirow}
\usepackage{diagbox}
\usepackage{subcaption}
\usepackage{subfloat}
\usepackage{xcolor}
\usepackage{lipsum}
\usepackage[font=footnotesize]{caption}
\mathchardef\mhyphen="2D

\title{\LARGE \bf
Point and Go: Intuitive Reference Frame Reallocation in Mode Switching for Assistive Robotics
}

\author{Allie Wang$^{*}$, Chen Jiang$^{*}$, Michael Przystupa$^{*}$, Justin Valentine$^{*}$, Martin Jagersand$^{*}$
\thanks{*Department of Computing Science, University of Alberta, Edmonton Alberta, Canada, T6G 2E8. {\tt\small \{luo3, cjiang2, przystup, jvalenti, mj7\}@ualberta.ca}}
}

\begin{document}

\maketitle
\thispagestyle{empty}
\pagestyle{empty}
\begin{abstract}
Operating high degree of freedom robots can be difficult for users of wheelchair mounted robotic manipulators. Mode switching in Cartesian space has several drawbacks such as unintuitive control reference frames, separate translation and orientation control, and limited movement capabilities that hinder performance. We propose Point and Go mode switching, which reallocates the Cartesian mode switching reference frames into a more intuitive action space comprised of new translation and rotation modes. We use a novel sweeping motion to point the gripper, which defines the new translation axis along the robot base frame's horizontal plane. This creates an intuitive `point and go' translation mode that allows the user to easily perform complex, human-like movements without switching control modes. The system's rotation mode combines position control with a refined end-effector oriented frame that provides precise and consistent robot actions in various end-effector poses. We verified its effectiveness through initial experiments, followed by a three-task user study that compared our method to Cartesian mode switching and a state of the art learning method. Results show that Point and Go mode switching reduced completion times by 31\%, pauses by 41\%, and mode switches by 33\%, while receiving significantly favorable responses in user surveys.\\

\end{abstract}

\section{Introduction}
In the USA alone, there are over 17 million people with an independent living disability and over 9 million people with a self-care disability \cite{c1}. Wheelchair-mounted robotic arms (WMRAs), such as the Manus and the Jaco, can reduce caregiving time and help these individuals regain autonomy by allowing them to independently perform activities of daily living (ADLs) \cite{c3, c4}. WMRA candidates are often used to operating input devices with only 2 to 3 degrees of freedom (DOF), which poses a challenge to use arms that have up to 7-DOF. Thus, improving WMRA control systems has the potential to enhance the quality of life of millions globally.

\setlength{\belowcaptionskip}{-15pt}
\begin{figure}[t]
\centering
\includegraphics[width=3.2in]{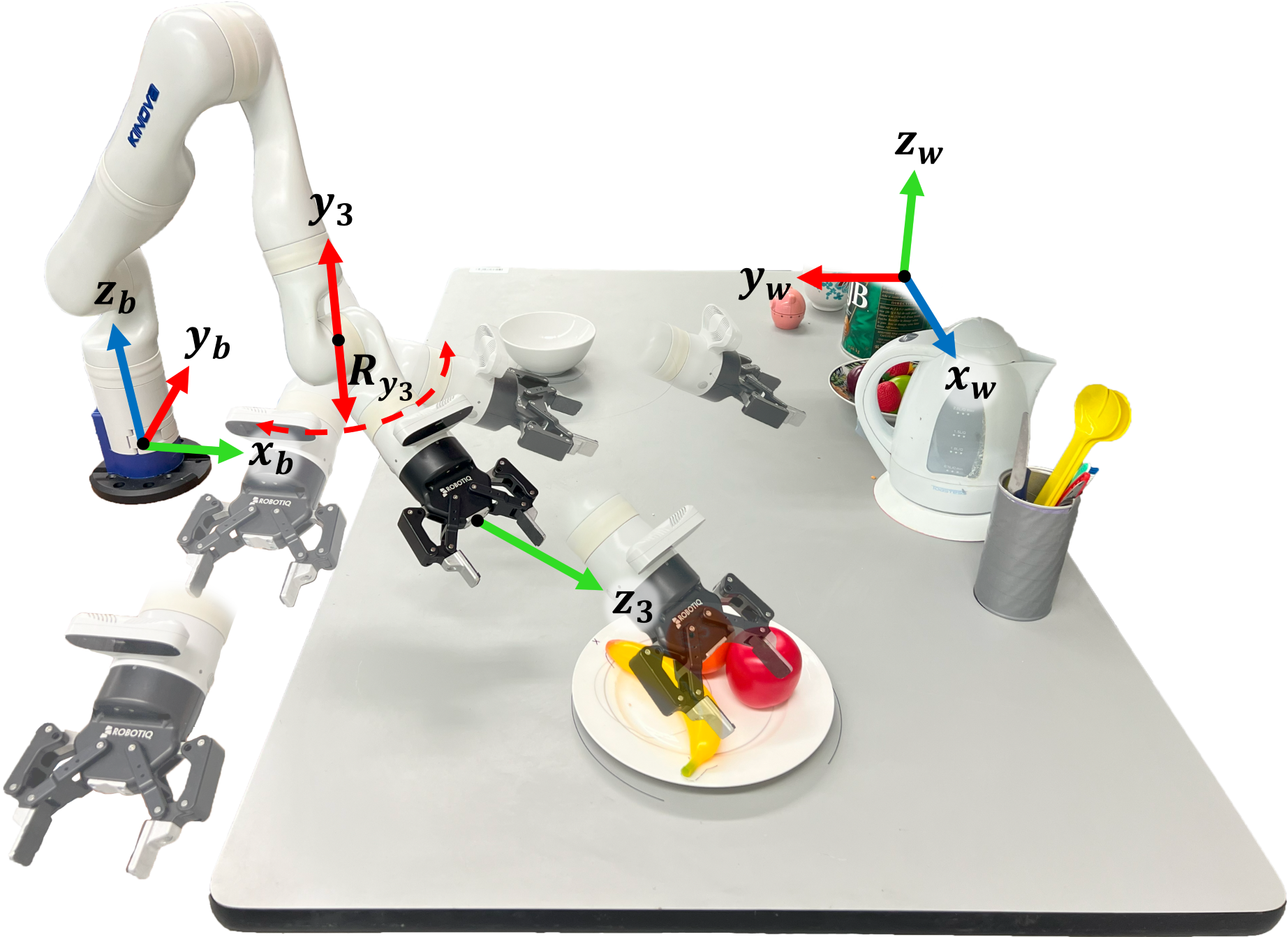}
\caption{Point and Go control axes consisting of wrist rotations about ${y_3}$ that `points' the end-effector to `go' along the axis defined by $z_3$.}
\label{coords}
\end{figure}
\setlength{\belowcaptionskip}{0pt}

Mode switching, the current control standard, allows users to switch between predefined subsets of movement axes. In practice, users commonly choose between a translation mode or a rotation mode which is combined with velocity control in which joystick inputs induce velocity impulses in the arm. These action spaces can exist in a Cartesian space, relative to the robot base frame, or in a pilot mode, based on the end-effector frame. Studies with the Manus arm showed that Cartesian space outperformed pilot mode \cite{c5}. However, despite Cartesian systems being the current control standard, recent studies have concluded that it places a high cognitive load on its users and imposes excess mode switches \cite{c7, c34, c35}.

In this paper, we address the fundamental issues of Cartesian mode switching. We propose a new system, Point and Go (PnG) mode switching, that can work \lq out of the box' for current WMRA users and can provide a superior framework for other methods to build on. Our translation mode provides efficient and intuitive control of both the position and orientation of the end-effector and its human-like wrist movements aim to enable more natural control strategies. Further, our rotation mode no longer requires intricate understanding of the end-effector frame and is combined with position control to provide consistent, intuitive and precise orientation adjustments of the end-effector. We highlight the following contributions in our Point and Go mode switching system:
\begin{itemize}
\item A new coordinate frame is created in the end-effector space for translations and rotations to improve consistency and lower cognitive load for the end user.
\item The end-effector is used as a physical fixture to define a dynamic translation axis along the base frame's horizontal plane. Lateral wrist motions orient this translation axis and enable complex movements to be performed in an intuitive and efficient manner.
\item Position control is integrated into the rotation mode to improve its consistency, ease of use, and precision.
\item We verify our results with initial ablation studies followed by a three-task user study comparing our method against Cartesian mode switching and learning-based State-Conditioned Linear Maps.
\end{itemize}

\section{Related Work}
Studies with the Manus evaluated pilot mode against Cartesian mode switching and found that pilot mode had 18\% slower task completion times and 23\% more mode switches \cite{c34}. Users reported challenges in tracking the gripper's coordinate frame in pilot mode, which increased both physical and cognitive load. More recent studies have concluded that Cartesian action spaces were confusing to operate and mode switching negatively impacts task completion time while increasing cognitive load \cite{c7, c35}. A study analyzing daily routines identified door opening as one of the most frequent tasks during activities of daily living \cite{c32}, and \cite{c33} observed active wrist flexion during door opening. Current mode switching systems lack the capability to replicate wrist motions, which consist of simultaneous translations and rotations. This may impede users' ability to carry out familiar movements and encourages excess mode switches, which has been shown to decrease user satisfaction \cite{c7}. 

Orientation adjustments with Cartesian Mode switching can be difficult to understand and control. For the same control input, subtly different end-effector poses can result in drastically different robot actions, requiring careful consideration of robot pose and complex input commands to complete tasks, resulting in increased mental load and decreased control efficiency. Campeau-Lecours et al. alleviated confusion in the rotation mode of these systems by introducing an adaptive $x$-axis of rotation that always lies on the horizontal plane. It was defined as the cross product between the base frame $z$-axis and the end-effector $z$-axis. However, the cross product operation is undefined at certain positions of the robot, which then requires a heuristic to define the reference frame in practice. Nevertheless, it was concluded that their method was more intuitive and efficient to operate \cite{c35}. 

Autonomous control systems have also been introduced to assist the user and reduce cognitive load. Supervisory methods prompt the user to provide high-level input for task goals, resulting in the system autonomously performing the task. These methods rarely conduct experiments beyond single-step tasks such as reach and grasps \cite{c20, c8, c9, c10, c11, c12, c13, c25, c26}. VGS implemented a larger array of tasks such as pouring and cabinet opening \cite{c27}, but required fiducial tags and many predefined waypoints, which is impractical in actual use. Some studies also show that more system autonomy does not always lead to greater user satisfaction \cite{c6, c29, c30, c31}.

Shared-autonomy methods collaboratively operate the robot with the user using predefined templates to transform control coordinate frames \cite{c28} or learning methods to find low-DOF representations of robot actions \cite{c21, c23, c24, c22}. These methods define task-specific routines, but defining suitable representations for all desired tasks could be intractable. Moreover, \cite{c28, c21, c23, c24} fail to consider how the user would switch between templates or sets of learned actions for different tasks in the same environment, which impedes real-world implementation. Mode switch assistance and user intent inference are other methods to reduce the cognitive load for users \cite{c7, c14, c15, c16, c17, c18, c19 }. These methods use ground truths such as user intent\cite{c7, c14}, or the set of possible task goals \cite{c15, c16, c17, c18, c19}, that are difficult to acquire in practice. Additionally, these methods are usually built on top of Cartesian mode switching, and thus inherit its drawbacks.

\section{Methods}
Although end-effector oriented action spaces were concluded to be unintuitive by past studies, we argue that they can be effective when simplified in an intuitive manner. In this section, we describe the methods used to achieve such a system that we call Point and Go (PnG) mode switching.

\subsection{Rotation Mode Coordinate Frame}
\begin{figure}
	\centering
  \setlength{\belowcaptionskip}{-10pt}
	\begin{subfigure}{0.51\linewidth}
		\includegraphics[width=\linewidth]{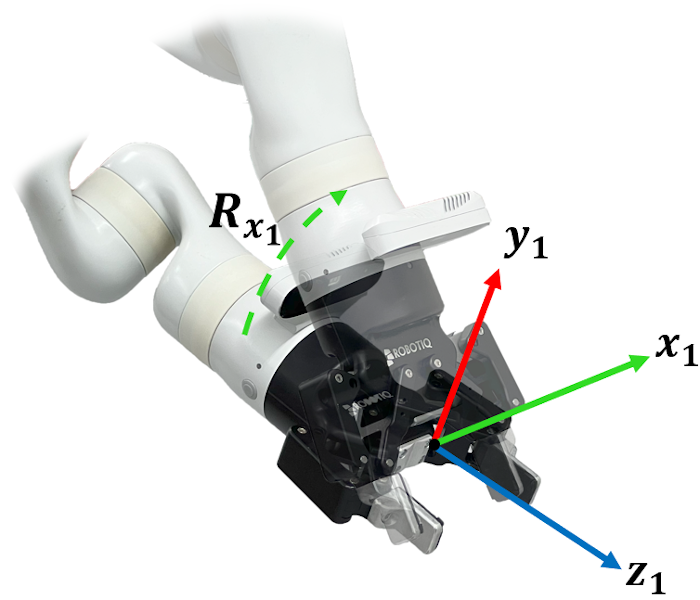}
		\caption{}
		\label{cartrot:a}
	\end{subfigure}
	\begin{subfigure}{0.47\linewidth}
		\includegraphics[width=\linewidth]{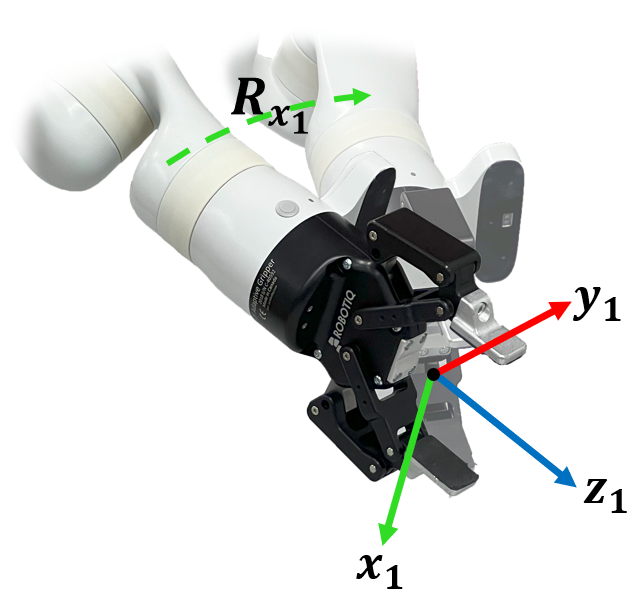}
		\caption{}
		\label{cartrot:b}
	\end{subfigure} 
 \setlength{\belowcaptionskip}{-15pt}
	\caption{Rotations along the $x_1$-axis for different end-effector orientations in Cartesian mode switching. Rotating the end-effector pose by 90 degrees between a) and b) causes different robot motions for the same joystick input in each case.}
	\label{cartrot}
\end{figure}

We define a new way to obtain the coordinate frame introduced by Campeau-Lecours et al. \cite{c35}, $[x_2, y_2, z_2]$, shown in Fig. \ref{theta}. We find it by rotating the end-effector frame, $[x_1, y_1, z_1]$, along the $z_1$-axis by $\theta_{align}$. The relation between $[x_2, y_2, z_2]$, $[x_1, y_1, z_1]$, and $\theta_{align}$ is shown in Fig. \ref{theta}. When compared to Cartesian rotations in Fig. \ref{cartrot}, a rotation in the $x_2$-axis remains consistent and intuitive in different orientations, thus reducing the mental load for the user as they will not have to closely monitor end-effector pose.

The user defines $\theta_{align}$ by joystick twists in the rotation mode. A PID controller then governs the last robot joint (joint 7, which is aligned with $z_1$) such that $x_2$ always lies parallel to the horizontal plane defined by $x_b\mhyphen y_b$, as shown in Fig. \ref{coords}. The controller minimizes the height difference between the $x_2$ vector and the end-effector origin in base coordinates. Defining the vectors in this way avoids the cross product operation and cone restriction in \cite{c35} and results in the same behavior as operating the last joint directly, like in the Cartesian system, but explicitly defines $\theta_{align}$ so that we can define joystick controls in this new coordinate frame.

\subsection{Point and Go Translations and Sweeping Motion}
In Cartesian control, the orientation of the WMRA base frame with respect to the environment affects how users complete a task. When the $x_b$ and $y_b$ axes are not in line with the workspace, combinations of inputs are required to move along the workspace axes, which may decrease intuitiveness and increase mental load. Reference frames such as the vertical axis, $z_b$, and the horizontal plane of the base frame, $x_b\mhyphen y_b$, do not suffer from this drawback and remain aligned with the workspace. Pilot mode defined the entire set of translations in the end-effector space, which was mentally taxing and difficult to keep track of \cite{c34}. By reallocating reference frames to be partially end-effector and base-frame oriented, we aim to create a control system that produces natural, intuitive motions, regardless of workspace alignment.

In our translation mode, translations are defined along two vectors: $z_b$, the base frame vertical axis controlled by joystick twists and $z_3$, the projection of $z_2$ onto the horizontal base plane controlled by joystick forward-back tilts. Translations along $z_b$ allow for interactions with different workspace heights. Translations along $z_3$ allow the robot to traverse the base horizontal plane towards where the end-effector is pointing.

The third DOF in our translation mode will be used to perform lateral wrist motions, as shown by ${R_{y_3}}$ in Fig. \ref{coords}, which changes the direction of the translations defined by $z_3$. The motion is defined by a rotation about the $y_3$-axis, located at the spherical center of the robot wrist, induced by joystick left-right tilts. This sweeping-like motion is performed by the spherical wrist joints 5, 6, and 7 and does not fix the end-effector position. This motion aims to emulate human wrist flexion and allows for natural movement strategies, such as pulling doors open with the wrist. 

By coupling horizontal base plane translations to the end-effector, we give the user a convenient fixture that explicitly defines the motion of the arm, putting the `Point' in Point and Go. With this method, curved motions that follow the end-effector can be performed to efficiently navigate the arm into target poses without having to switch modes. The system also retains the ability to perform small lateral translations by the small angle approximation. The physical pointer of the end-effector aims to provide an effective indicator through which users can traverse the horizontal task space with reduced cognitive load compared to other systems.

\subsection{Position Control for Rotation Mode}
In the rotation mode of the arm, we define rotations in the $[x_2, y_2, z_2]$ frame, which allows for consistent and intuitive rotation control that does not depend on end-effector pose. We map pitch ($R_{x_2}$; joystick forward-back tilts) and yaw ($R_{y_2}$; joystick left-right tilts) to position control, managed by a PID controller. When switching into rotation mode, the current end-effector pose is defined as the home pose. User inputs define a goal vector that is rotationally displaced about the $x_2$ and $y_2$ axes, with a maximal angular displacement of $\pm\alpha$ with respect to the home pose. The PID controller then performs orientation adjustments to mirror this goal vector. Fig. \ref{theta} shows the goal vector in orange when the user inputs a joystick deflection in $+x_2$.

Position control allows the user to perform more precise adjustments that are not influenced by joystick impulses as in velocity control. Instead, joystick deflections defined by the user are mirrored by the robot. Undoing mistakes with this mode is simple: the robot will return to the home pose when the joystick is in its neutral position. When the pose has been adjusted to a desired position, exiting rotation mode sets the new orientation of the end-effector. Orientation changes outside the scope of $\alpha$ can be achieved by reinitializing the home position by exiting and reentering rotation mode. Though this may increase the number of mode switches, we believe that position control offers greater precision and ability to undo mistakes, leading to better overall performance. By combining this position controller with the intuitive orientation changes in the $[x_2, y_2, z_2]$ frame, we create an optimized rotation control mode that is consistent, precise, and intuitive.

\setlength{\belowcaptionskip}{0pt}

\begin{figure}
	\centering
 \setlength{\belowcaptionskip}{-10pt}
	\begin{subfigure}{0.49\linewidth}
		\includegraphics[width=\linewidth]{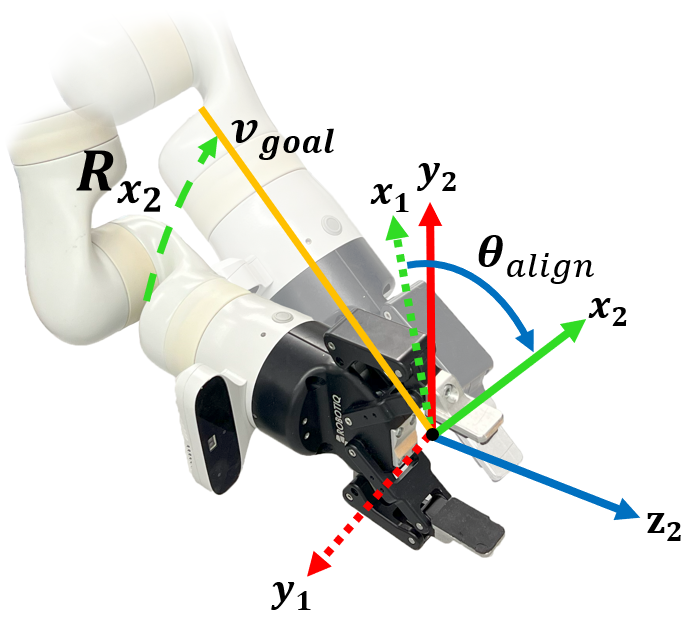}
		\caption{}
		\label{theta:a}
	\end{subfigure} 
    \begin{subfigure}{0.49\linewidth}
        \includegraphics[width=\linewidth]{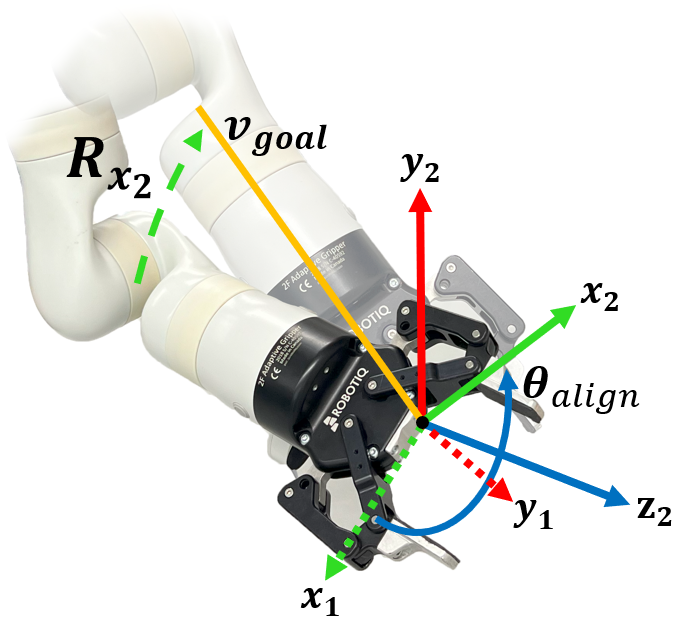}
        \caption{}
        \label{theta:b}
    \end{subfigure} 
\setlength{\belowcaptionskip}{-20pt}
	\caption{Rotations along the $x_2$-axis for different end-effector orientations in the new coordinate frame. Robot motions stay consistent regardless of end-effector pose, leading to more intuitive control}
	\label{theta}
\end{figure}
\setlength{\belowcaptionskip}{0pt}

\section{Experiments}
A rotation control experiment and a translation control experiment were separately conducted to evaluate PnG's additions in isolation. A user study was then conducted that compared PnG mode switching to Cartesian mode switching and State-Conditioned Linear Maps (SCL) from Przystupa et al., which uses deep learning to find a 2-DOF linear representation of high dimensional robot actions given the robot's state \cite{c23}. For all studies, users operated a WMRA system with a 7-DOF Kinova Gen3 robot and a 3-DOF joystick. Three buttons on the joystick were used to change modes and open and close the gripper. All control systems were limited to the same end-effector velocities. For PnG, $\alpha$ was set to 45$^{\circ}$.

These experiments were approved by University of Alberta Research Ethics and Managements (Pro00054665) and signed consent was obtained from the 11 non-disabled participants (ages 22-28, 6 female) in this study. Each participant used all control systems to complete all tasks. Experiments followed a single-blind format with the control system order randomized. Prior to each experiment, tutorials were provided for each control system followed by one practice trial.

\subsection{Rotation Control Study and Translation Control Study}
A rotation control study was conducted to compare the performance of position control to velocity control in the new rotation mode reference frame. These two systems were also compared to the baseline of the rotation mode in Cartesian mode switching. In the experiment, users were tasked with aligning a laser pointer in the robot hand with a set of coloured circle targets, shown in Fig. \ref{exp:r}. Completion time of five trials were recorded per control scheme. Each trial consisted of moving the pointer from the starting position, with noise added to joint 7, to a static position over the target. Users were asked to rank each control system in order of preference at the end of the experiment.

The translation study was conducted to compare the effectiveness of PnG's translation mode to the baseline of Cartesian mode switching. PnG was evaluated in this experiment without its rotation mode accessible to the user to determine if wrist motions and base plane translations were sufficient to translate and orient the robot without switching modes. A goalpost, requiring specific end-effector orientations to pass through it, was placed on the workspace, as shown in Fig. \ref{exp:t}. Two designated goalpost positions were evaluated. From a start position with noise added to joints 5 and 7, the user was tasked with moving the end-effector through the goal posts. Completion time of two trials was recorded for each goal position. Users were asked which translation system they preferred at the end of all trials.

\subsection{User Study}
After the initial studies, a user study was conducted to compare performance of PnG to Cartesian mode switching and SCL. The testing setup is shown in Fig. 4c-f, where users were seated in a kitchen workspace.

\subsubsection{Study Protocol and Metrics}
\begin{figure*}[t]\vspace{0.2cm}
	\begin{subfigure}{0.33\columnwidth}
		\includegraphics[width=\linewidth]{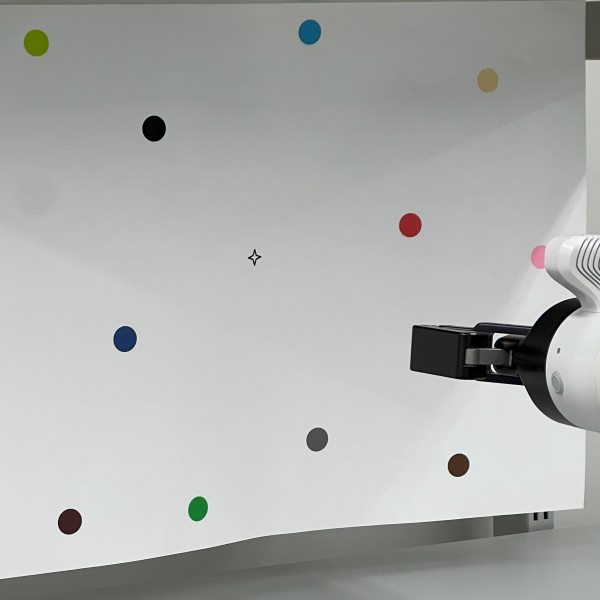}
		\caption{Rotation experiment}
		\label{exp:r}
	\end{subfigure} 
        \begin{subfigure}{0.33\columnwidth}
        \includegraphics[width=\linewidth]{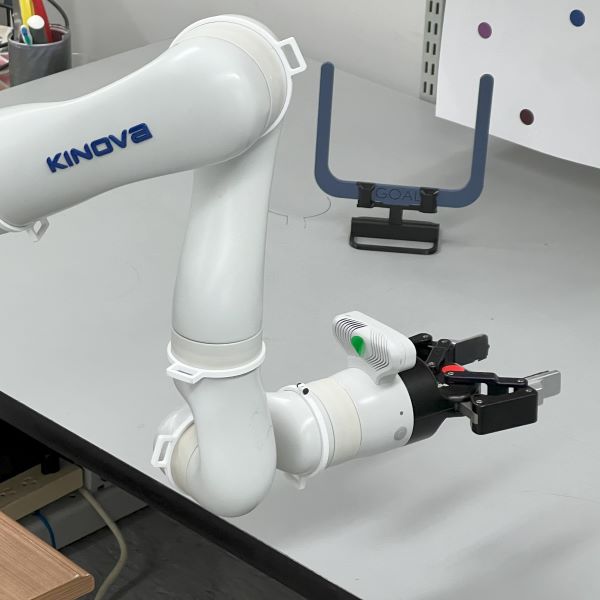}
        \caption{Translation experiment}
        \label{exp:t}
        \end{subfigure} 
	\centering
	\begin{subfigure}{0.33\columnwidth}
		\includegraphics[width=\linewidth]{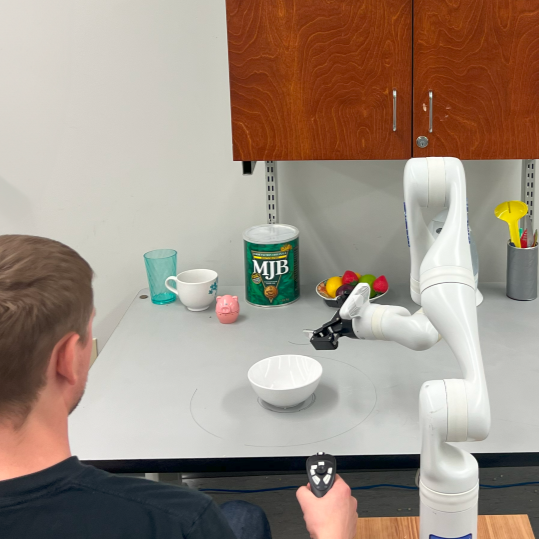}
		\caption{User study start position}
		\label{exp:setup}
	\end{subfigure}
	\begin{subfigure}{0.33\columnwidth}
		\includegraphics[width=\linewidth]{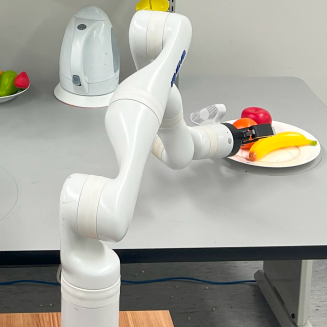}
		\caption{Banana Pick and Place}
		\label{exp:banana}
	\end{subfigure}
	\begin{subfigure}{0.33\columnwidth}
        \includegraphics[width=\linewidth]{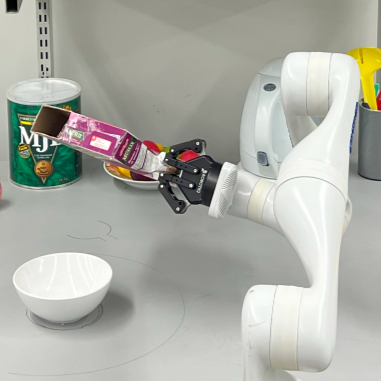}
        \caption{Cereal Pick and Pour}
        \label{exp:pour}
        \end{subfigure}
        \begin{subfigure}{0.33\columnwidth}
        \includegraphics[width=\linewidth]{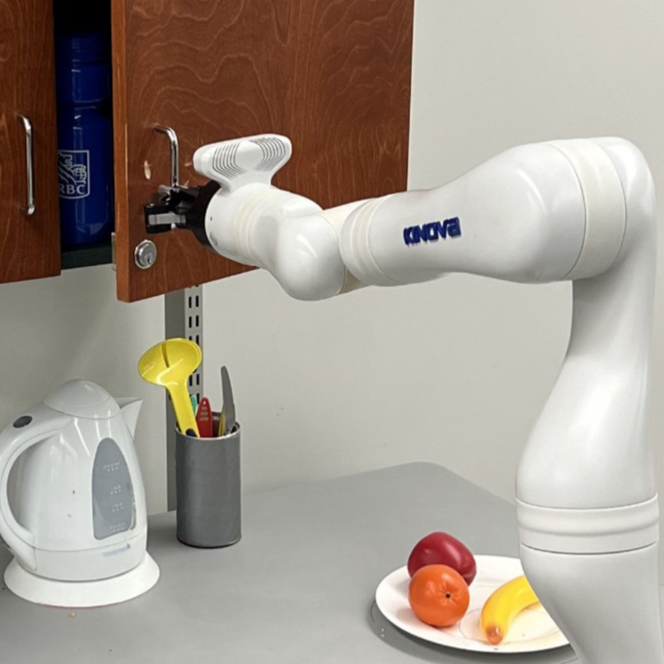}
        \caption{Cabinet Opening}
        \label{exp:cabinet}
         \end{subfigure}     
\setlength{\belowcaptionskip}{-18pt}
	\caption{Experimental setup and tasks for the ablation experiments and user study.}
	\label{exp}
\end{figure*}

Participants were asked to perform three tasks with the robot. For each task and control system, two trials were recorded. The same robot starting position, shown in Fig. \ref{exp:setup}, was used for all tasks with an added 20$^{\circ}$, 20$^{\circ}$ and 360$^{\circ}$ of noise to the last three joints, respectively. At the end of each task, a survey consisting of a NASA Task Load Index (NASA-TLX) assessment\cite{c36} and a 7 point Likert survey was administered as follows:

\begin{itemize}
    \item \textit{Control}: How in control of the system did you feel?
    \item \textit{Intuitive}: How intuitive was the system to use? 
    \item \textit{Undo}: How easy was it to undo mistakes in the system?
    \item \textit{Precise}: How precise were commands in the system?
    \item \textit{Prefer}: Do you prefer to use this system for this task?
\end{itemize}

Videos, joystick inputs, joint positions and joint velocities were recorded. From this data, task completion time, mode switches, and pause count were calculated for each trial. Mode switches and pauses were not calculated for SCL, as it did not have modes to switch between. Mode switches were defined as any time the user changed control modes and pauses were defined as any discontinuity in joystick input. 

Comparisons were achieved using a two-way repeated measures ANOVA. Sphericity was assessed with Mauchly’s test of sphericity. A post-hoc pairwise t-test was used to find the effect of the control systems within each experiment. A Wilcoxon signed-rank test was used to evaluate Likert scale results and other data if normality was violated. The threshold for significance for all tests was 0.05.

\setlength{\belowcaptionskip}{0pt}

\subsubsection{Tasks}\hfill\\
\indent\textbf{Banana Pick and Place:} The user had to pick up a banana from a plate on one end of the table and bring it to the target bowl, as shown in Fig. \ref{exp:banana}. Two other fruits on the plate blocked access to the banana from certain angles.

\textbf{Cereal Pick and Pour:} The user had to grasp a cereal box from an open overhead cabinet and pour its contents into the target bowl, as shown in Fig. \ref{exp:pour}. The cereal box was only open on one end, requiring the user to pour the box in certain orientations.

\textbf{Cabinet Opening:} The user had to open a cabinet door 90$^{\circ}$ or greater with respect to the closed position, as shown in Fig. \ref{exp:cabinet}. Participants were shown the position the door would remain open and could then be pushed open from the inside.

\section{Results}
\subsection{Rotation Control Study and Translation Control Study}
Results for the rotation study are shown in the first three columns of Table \ref{results0}. Users took an average of 10.06 s with the Cartesian mode switching baseline. When using the new orientation frame, velocity control resulted in an average time of 5.93 s while position control averaged a time of 4.51 s. There was a significant difference between all three of these values. Further, 55\% of participants preferred position control while 36\% preferred velocity control.

Results from the translation study are shown in the last two columns of Table \ref{results0}. Point and Go took an average of 12.45 seconds to reach the target, while the Cartesian mode switching baseline took an average of 33.56 seconds. All users indicated PnG as their preferred control system.

\subsection{User Study}
Table \ref{results1} shows the completion time, NASA-TLX Workload, mode switches and pauses observed in the user study. Mode switches and pauses are only reported for PnG and Cartesian mode switching. Fig. \ref{likert} shows median Likert survey responses. Across all experiments, significant differences in completion time was found between Cartesian mode switching's time of 111.9 s and SCL's time of 126.8 s with respect to PnG's time of 76.8 s. Additionally, PnG had an overall workload rating of 45.9, significantly less than SCL's workload of 57.2. Cartesian mode switching's 52.7 pauses and 11.1 mode switches was also significantly greater than PnG's 31.2 pauses and 7.4 mode switches. Median survey results across all experiments found significant differences in favour of PnG when compared to the Cartesian system's preference, control, intuitiveness, and precision ratings and in all five of SCL's ratings.

\textbf{Banana Pick and Place:} We observed significantly better results for PnG in completion time, mode switch count, pause count and workload score when compared to the Cartesian baseline. On average, PnG completed the task 44.0 seconds faster, with 5.4 fewer mode switches and 21.3 fewer pauses per trial and received significantly more favorable responses in preference and control ratings compared to the Cartesian system. No significant differences were found between PnG and SCL in the survey responses or quantitative metrics.

\textbf{Cereal Pick and Pour:} PnG's completion time of 107.0 s was significantly faster than the Cartesian system's 135.9 s or SCL's 145.8 s. Compared to SCL, PnG's workload was significantly less by 13.2. Compared to the Cartesian baseline, PnG had a significant difference of 26.9 fewer pauses. Likert surveys resulted in significantly favorable control and preference ratings for PnG compared to the other two systems. In addition, PnG received further significant differences in precision, intuitiveness and ability to undo mistakes when compared to SCL.
\setlength{\textfloatsep}{0.3cm}
\begin{table}[b]
\renewcommand{\arraystretch}{1.15}
\caption{Quantitative results from the ablation experiments.}
\label{results0}
\centering
\begin{footnotesize}
\setlength{\tabcolsep}{3pt}
\resizebox{\columnwidth}{!}{%
\begin{tabular}{c|ccc|cc}
\hline
& \multicolumn{3}{c|}{Rotation}& \multicolumn{2}{c}{Translation}\\
\cline{2-6}
& Pos. ctrl & Vel. ctrl & Cart. & PnG & Cart.\\
\hline
Time (s) & 4.51$\pm$0.25 & 5.93$\pm$0.38 &10.06$\pm$0.83 &12.45$\pm$0.62&33.56$\pm$2.47\\
Pref. & 6/11 & 4/11 & 1/11&11/11&0/11\\
P-value & - & 0.029 & P\textless0.001 &-&P\textless0.001\\
\hline
\end{tabular}}
\end{footnotesize}
\end{table}

\textbf{Cabinet Opening:} Significantly better mean completion time, mode switch count and pause count was observed for the PnG system compared to the Cartesian baseline. On average, PnG completed trials 32.6 s faster, with 3.2 fewer mode switches and 16.3 fewer pauses. When compared to SCL, significant differences in completion time and workload were found, with PnG completing trials 123.1 s faster and with a 20.5 lower workload rating. Likert surveys showed significantly more favorable responses in preference, control, and precision for PnG compared to the other two systems. Further, PnG showed significantly improved ratings in intuitiveness and the ability to undo mistakes when compared to the SCL baseline.

\setlength{\belowcaptionskip}{5pt}
\begin{table*}[t]
\renewcommand{\arraystretch}{1.2}
\caption{Quantitative results$^{1}$ from the user study comparing control systems for three tasks.}
\label{results1}
\centering
\begin{footnotesize}
\setlength{\tabcolsep}{3.9pt}
\begin{tabular}{c|ccc|ccc|cc|cc}
\hline
\multirow{2}{*}{\diagbox{{Experiment}}{{Metric}}} & \multicolumn{3}{c|}{Completion Time} & \multicolumn{3}{c|}{NASA-TLX Workload}& \multicolumn{2}{c|}{Mode Switches} & \multicolumn{2}{c}{Pauses}\\
\cline{2-11}
& PnG & Cart. & SCL & PnG & Cart. & SCL & PnG & Cart. & PnG & Cart. \\
\hline
Banana Pick and Place & \bf{62.9$\pm$6.5} & 106.9$\pm$9.8 & 51.2$\pm$3.8 & \bf{43.5$\pm$3.1} & 51.5$\pm$3.8 & 43.7$\pm$4.6&  \bf{4.1$\pm$0.9} & 9.5$\pm$1.1 & \bf{25.1$\pm$3.6} & 46.4$\pm$6.5\\
Cereal Pick and Pour & \bf{\underline{107.0$\pm$10.6}} & 135.9$\pm$9.3 & 145.8$\pm$14.6 & \underline{51.3$\pm$ 3.6} & 55.5$\pm$4.5 & 64.5$\pm$4.5&  14.6$\pm$2.0 & 17.1$\pm$1.9 &  \bf{44.5$\pm$4.4} & 71.4$\pm$7.9 \\
Cabinet Opening & \bf{\underline{60.4$\pm$2.4}} & 93.0$\pm$9.1 & 183.5$\pm$14.6 & \underline{42.8$\pm$ 2.7} & 49.3$\pm$ 5.5 & 63.3$\pm$5.0& \bf{3.5$\pm$0.7} & 6.7$\pm$0.7& \bf{23.9$\pm$2.2} & 40.2$\pm$4.7 \\
Overall & \bf{\underline{76.8$\pm$5.0}} & 111.9$\pm$5.8 & 126.8$\pm$10.0 & \underline{45.9$\pm$1.9} & 52.1$\pm$2.6 & 57.2$\pm$3.1 & \bf{7.4$\pm$1.0} & 11.1$\pm$0.9  & \bf{31.2$\pm$2.4} & 52.7$\pm$4.1 \\
\hline
\multicolumn{11}{l}{\scriptsize$^{1}$ Results are reported with standard error. Bolded and underlined values indicate PnG was significantly different than Cartesian and SCL systems, respectively.} \\
\end{tabular}
\end{footnotesize}
\end{table*}

\setlength{\belowcaptionskip}{-3pt}
\begin{figure*}[t]\vspace{-0.3cm}
	\centering
	\begin{subfigure}{0.5\columnwidth}
		\includegraphics[width=\linewidth]{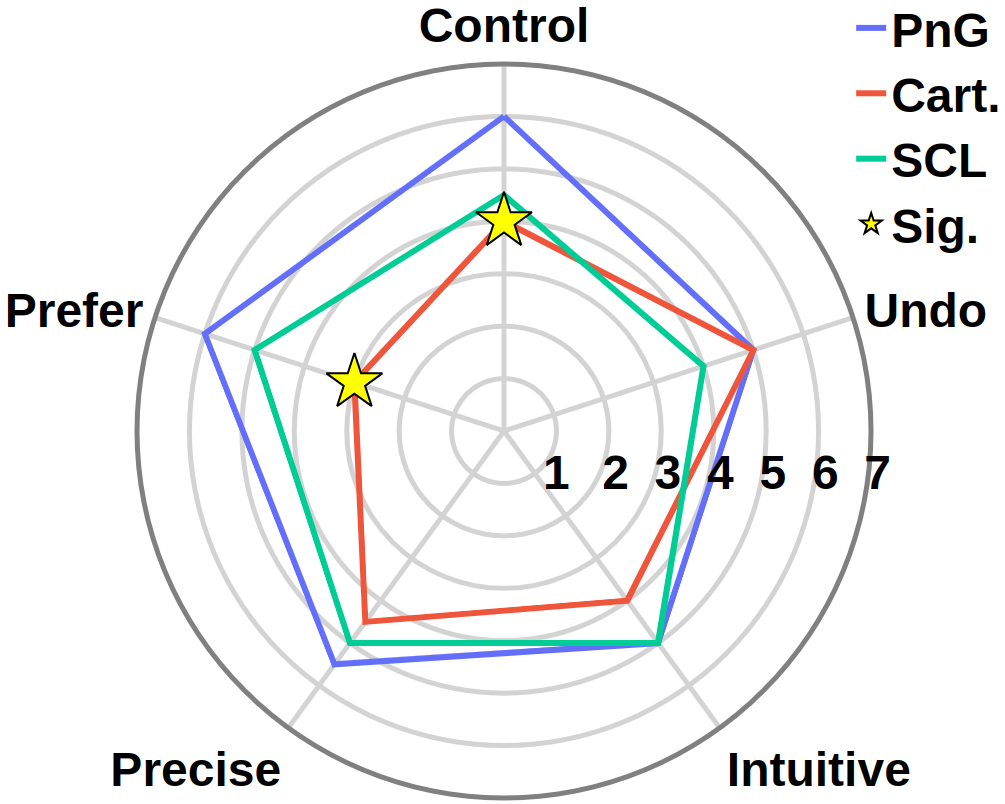}
		\caption{Banana Pick and Place}
		\label{likert:ban}
	\end{subfigure}
	\begin{subfigure}{0.5\columnwidth}
		\includegraphics[width=\linewidth]{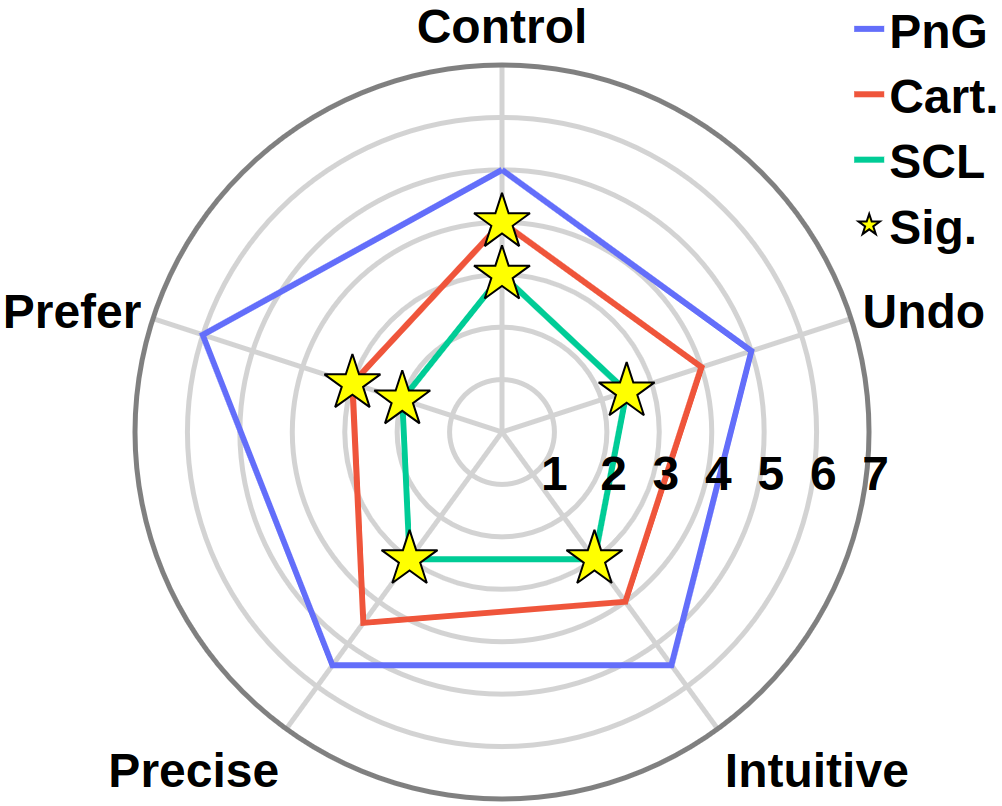}
		\caption{Cereal Pick and Pour}
		\label{likert:pou}
	\end{subfigure}
	\begin{subfigure}{0.5\columnwidth}
        \includegraphics[width=\linewidth]{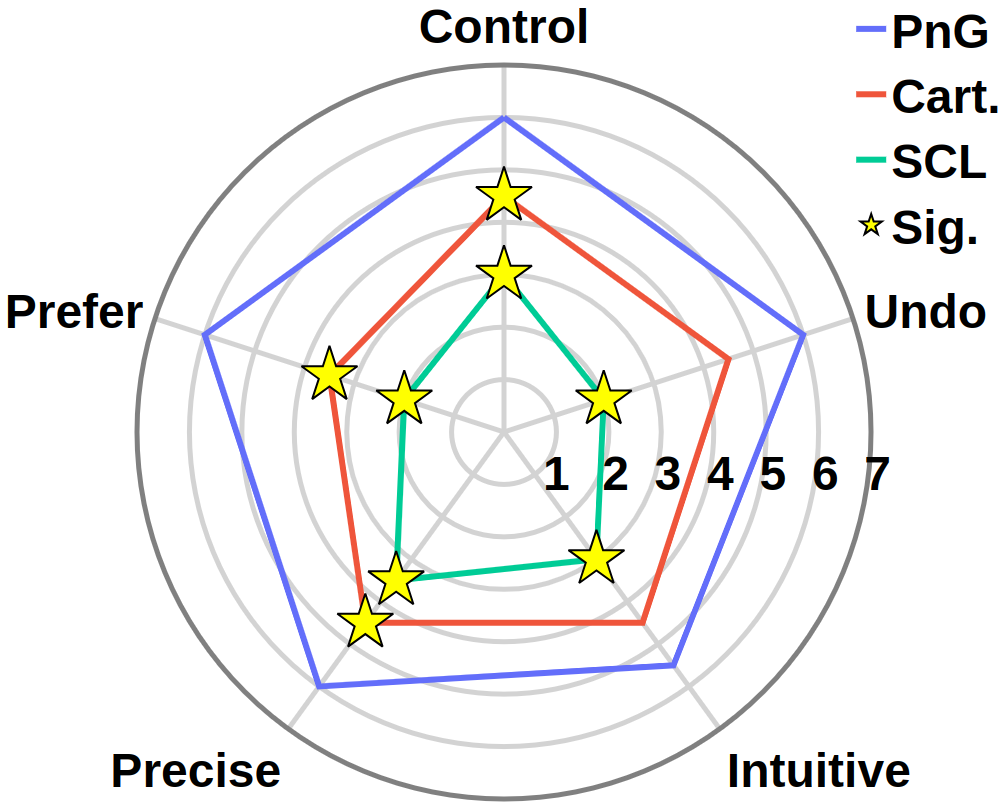}
        \caption{Cabinet Opening}
        \label{likert:cab}
     \end{subfigure}
    \begin{subfigure}{0.5\columnwidth}
        \includegraphics[width=\linewidth]{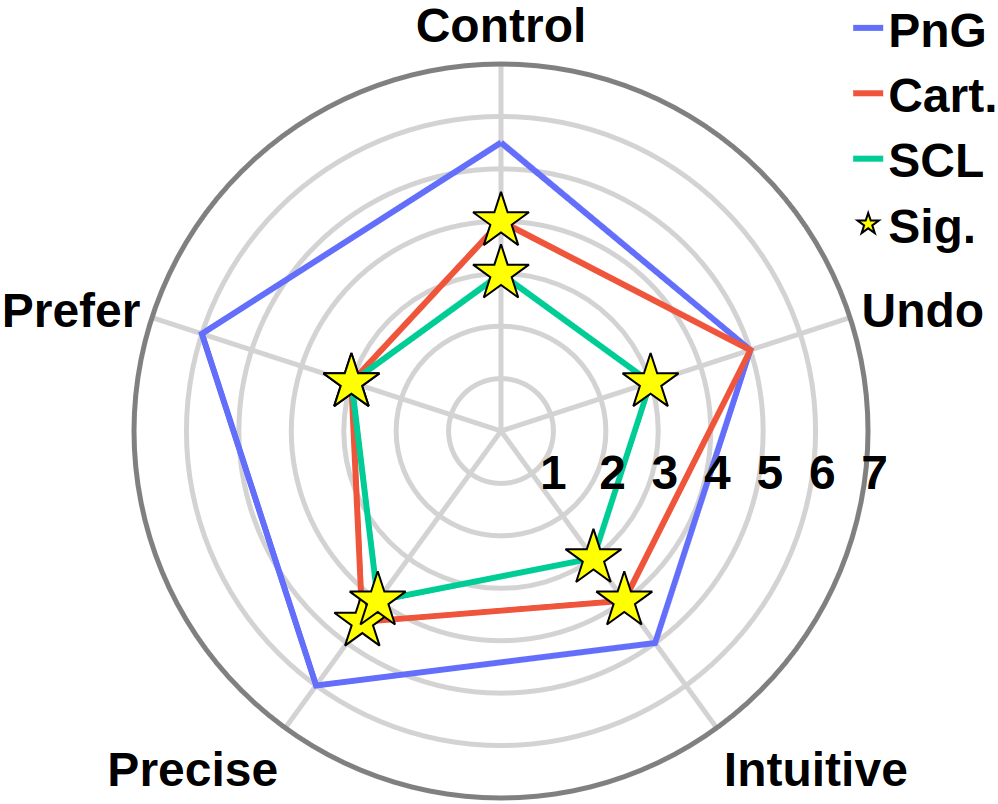}
        \caption{Across All Experiments}
        \label{likert:all}
     \end{subfigure}   
 \setlength{\belowcaptionskip}{-15pt}
	\caption{Radar plots showing median Likert survey responses (1-least favorable, 7-most favorable)}
	\label{likert}
\end{figure*}

\section{Discussion}
\subsection{Rotation Control Study and Translation Control Study}
The rotation study results show a marked improvement in user preference and completion time when using the new coordinate frame for end-effector rotations. The difference in completion times between velocity control and position control in the new coordinate frame was less pronounced but still significant. The 1.42 second improvement that position control achieved over velocity control points to users having an easier time completing precise alignments with this system. This is also reflected by the user preferences, with the majority of users preferring to use position control to perform orientation adjustments. However, it is important to note that for this experiment especially, the larger size of the joystick and physical capabilities of the user may skew these results in favour of position control.

When evaluating the performance of PnG translations to Cartesian mode switching, a pronounced improvement was observed. PnG, when restricted to just one mode, outperformed the full suite of Cartesian mode switching. We show that a smaller set yet optimized control axes may be easier and more intuitive for the end user to translate and roughly orient the end-effector towards goals all without switching modes. Despite Cartesian mode switching providing more degrees of control to the user, it resulted in drastically slower completion times during this experiment. It was especially challenging for users to operate the rotation mode in Cartesian mode switching to orient the end-effector. All users preferred PnG for this task, which points to the effectiveness of its translation mode that allows for intuitive control over both end-effector position and orientation during tasks.

\subsection{User Study}
\textbf{Banana Pick and Place:}
The shape of the banana necessitated specific end-effector poses to grasp it. We attribute the faster completion times and fewer mode switches of PnG, when compared to Cartesian mode switching, to its efficient and intuitive translations and rotations that allow for quick and precise control to achieve desired poses. Participants operating the robot using PnG mode switching had an easier time both navigating to the banana and orienting the gripper to grasp it. Further, fewer pauses may indicate less planning is required when using PnG. Under Cartesian control, users often struggled to find the desired inputs, as evidenced by the greater mode switches and pauses. Survey results supported this, as users reported lower workloads and responded more favorably in feelings of preference and control when using PnG. SCL, on the other hand, performed similarly to PnG in this experiment, with no significant difference in any recorded metric. Users efficiently completed the task with either of these two systems.

\textbf{Cereal Pick and Pour:}
Users perceived this task to be the most difficult, as it had the highest workload ratings among the three tasks. Precise control of translations and orientation adjustments were required to both grasp the box and pour out its contents. When grasping objects for pouring, the end-effector oriented rotation frame was no longer relevant to the task. Thus, attempts to pour the cereal often misaligned the spout, requiring users to repeatedly change between modes to correct for this. This issue occurred in both PnG and Cartesian systems. Thus, resulting observations may be a function of task difficulty and the challenge of orienting the end-effector in the two systems. This is reflected in the smaller differences in mode switch count, workload, and Likert survey responses between the two control systems.

PnG mode switching still outperformed Cartesian mode switching in completion time and pause count. But, for some users, we observed that the lack of lateral translations in the PnG system was restricting. Large lateral adjustments would require additional translation corrections along $z_3$ as well, which may have increased the mental load for some. This is reflected by the closer user control ratings between the two systems. Despite this, PnG was still overwhelmingly preferred over Cartesian mode switching by users.

SCL performed poorly for this task. Users often missed crucial alignments to required to complete the task, and found it difficult to perform the necessary corrections. We hypothesize that due to the complexity and large breadth of the task, simplifying it down to 2-DOF prevented the users from having full control authority, and increased dissatisfaction. This was reflected by the poor quantitative and qualitative measurements recorded for SCL, with PnG outperforming it in all metrics.

\textbf{Cabinet Opening:}
We can attribute the superior performance of PnG in completion time, mode switches and pauses, when compared to Cartesian mode switching, to its translation mode that allows users to easily perform complex translations and lateral rotations without switching modes. With PnG's sweeping wrist motions, users were able to open the cabinet, navigate the end-effector inside the gap of the door, and pull it open akin to how a human would open a slightly ajar door. We believe that users preferred the more natural movement capabilities our method provided, as supported by favorable survey responses. Since it was difficult to open the cabinet enough such that the end-effector could fit in the gap without changing orientation, Cartesian control required users to perform more adjustments in the rotation mode to manipulate the door, which contributed to increased pauses, mode switches, and completion times.

Like in cereal pouring, when the task requires complex and precise movements in many axes, users seem to struggle to perform the task with SCL. Aligning the end-effector with the cabinet handle and then pulling it about the hinge axis proved particularly difficult for users. This was again reflected by the poor quantitative and qualitative results observed in SCL, with PnG outperforming it in all metrics.

\textbf{Across All Experiments:}
Overall, we observed a significant reduction in task completion time, pauses and mode switches when using our system compared to Cartesian mode switching. These metrics support a reduction in cognitive load for the user when using PnG, indicating that it is a more intuitive control system that requires less planning during task execution. Overall, PnG reduced completion times by 31\%, workload by 12\%, mode switches by 33\%, and pauses by 41\% when compared to Cartesian mode switching, and reduced completion times by 39\% and workload by 20\% when compared to SCL. The benefits of PnG are further evidenced by the Likert responses across all experiments, where it achieved superior ratings in all categories except for undo-ability. Our results show that PnG mode switching is a promising control system that could replace Cartesian mode switching in WMRAs. Moreover, its performance matches that of a state of the art learning method in basic tasks, and exceeds it in complex tasks, all while working \lq out of the box', requiring no training, environment adjustments, ground truths or external sensors. Our results also support the claim that increased system autonomy, as seen in SCL, does not always improve user satisfaction.

\section{Conclusion}
In this paper, we propose PnG mode switching which defines new action subspaces that blend base frame and end-effector aligned control axes. Our system's translations use the end-effector as a simple guide to traverse the horizontal base plane. Its novel sweeping motions can also control end-effector orientation and allow for more natural interactions with the environment without switching modes. The position controlled rotation mode uses realigned reference frames to allow for consistent, precise and intuitive orientation adjustments independent of end-effector pose. Our experiments demonstrated that our method outperformed Cartesian mode switching through various quantitative measures and subjective survey responses, as well as matching or exceeding a state of the art learning method in SCL. Future work includes adapting control reference frames to grasped object interest points, such as the spout of a jug or a hinge of a door, to create more task-relevant control schemes to streamline complex tasks. Additionally, evaluating PnG in an at-home WMRA setting with experienced users would be useful to gain insight on the its real-world performance.
\addtolength{\textheight}{-1.615cm}  
\bibliographystyle{IEEEtran}

\bibliography{bibliography}

\end{document}